\documentclass[runningheads]{llncs}
\usepackage{graphicx}
\usepackage{xcolor}
\usepackage{framed}
\usepackage{listings}
\usepackage{multirow}
\usepackage{makecell}
\usepackage{comment}
\usepackage{graphicx}
\usepackage{amsmath}
\usepackage{float}
\usepackage{pifont}
\newcommand{\cmark}{\ding{51}}%
\newcommand{\xmark}{\ding{55}}
\begin{document}
\title{Abstract Operations Research Modeling Using Natural Language Inputs}
\titlerunning{NL2OR}
\author{
Junxuan Li\inst{1}\thanks{Corresponding author}\and
Ryan Wickman\inst{1} \and
Sahil Bhatnagar\inst{1}\and
Raj Kumar Maity\inst{1}\and
Arko Mukherjee \inst{1}}
\authorrunning{}
\authorrunning{Li, Wickman, Bhatnagar, Maity and Mukherjee}
%
\institute{Microsoft, Redmond, WA 98052 USA\\ \email{\{junxuanli,sabhatn,ramaity,ryanwickman, arko.mukherjee\}@microsoft.com} }
\maketitle              
\begin{abstract}
Operations research (OR) uses mathematical models to enhance decision-making, but developing these models requires expert knowledge and can be time-consuming. Automated mathematical programming (AMP) has emerged to simplify this process, but existing systems have limitations. This paper introduces a novel methodology that uses recent advances in Large Language Model (LLM) to create and edit OR solutions from non-expert user queries expressed using Natural Language. This reduces the need for domain expertise and the time to formulate a problem. The paper presents an end-to-end pipeline, named NL2OR, that generates solutions to OR problems from natural language input, and shares experimental results on several important OR problems.
\end{abstract}
\section{Introduction}
Operations research (OR) is a field that has been widely used to enhance decision-making by creating and using mathematical models of real-world situations to find solutions \cite{HillLieb01}. A major stream of OR solution is to use mathematical programs (MP) which consist mathematical representations of decision variables, optimization objectives, and decision constraints. For instance, in the area of e-commerce order fulfillment, a resource allocation MP aims to minimize total fulfillment cost, utilizing limited inventories located across the merchant's warehouse/retailing network. The feasible fulfillment decisions are required to satisfy operational constraints, e.g. channel restrictions, region restrictions, logistic restrictions, and etc.. Another OR problem example, in the area of resource scheduling optimization, a scheduling planner aims to optimize the utilization of human agent resources to accomplish tasks with skill set specifications. The feasible scheduling decisions are required to satisfy operational constraints, e.g. staffing requirements, shift scheduling, skill matching and etc.. For physical tasks, the scheduling planner will also need to minimize agent traveling, and making sure agent traveling schedules satisfy time windows for both task reservations and agent shifts. In both examples, the MP aims to optimize a certain objective function, subject to a set of operational constraints.

However, in the related industries, modeling an OR problem and developing an MP solution will require specialized knowledge set (e.g. mathematical modeling and operations research algorithm design training). For instance, in the e-commerce order fulfillment example, one needs to have knowledge in inventory optimization, transportation optimization, and logistics optimization. 
Therefore, the development of MP solutions is often left to a small group of domain experts. Moreover, developing and deploying a computer-based solution may take months. This long development cycle can be a barrier to OR solution adoption, especially in the context of small and medium-sized enterprises. 

To overcome this barrier, the field of automated mathematical programming (AMP) has emerged, aiming to automate the process of developing MP solutions. AMP aims to enable non-experts to develop and deploy MP solutions with little or no human intervention. AMP systems can be divided into two categories: \textit{closed-box} and \textit{open-box} systems.

\textit{Closed-box} AMP systems use machine learning techniques to learn the mapping between the input data and the optimal solution \cite{ZHANG2023205,Nair2021}. These systems require no prior knowledge of the mathematical structures. However, the solutions produced by black-box systems are often sub-optimal and cannot be guaranteed to satisfy operational constraints.

\textit{Open-box} AMP systems, on the other hand, are based on mathematical programming techniques and require the user to specify the mathematical model of the problem \cite{LiMellou2023,ramamonjison2022,Yang2023}. These systems start from detailed business logic and a problem data set, and generate a solver-specific script (also known as a concrete model) for user to review. After the generated model being reviewed and validated, it is able to guarantee that the solution satisfies the operational constraints, and can \textit{often} produce optimal solutions. However, the user needs to have a good understanding of the mathematical programming problem and how to interface with a variety of solvers/libraries (e.g. Gurobi, CPLEX, ORtools, AMPL).

To mitigate the limitations these drawbacks of existing closed- and open-box AMP systems, we propose a novel methodology that create new OR solutions and edit existing OR solutions using natural language (NL) queries. Recent advances in natural language processing (NLP) have led to the development of large language models (LLMs) that can be used for human language text generation. These models have been used for various tasks, such as language translation, question answering, and text summarization \cite{devlin2018bert,zhang2022opt,chowdhery2023palm,brown2020language,achiam2023gpt,touvron2023llama}. We propose to use LLMs to generate markup language passages for OR problems, which can then be converted into programming language passages. This technique entails obtaining user input for an OR problem, formulating an NL prompt grounded in the user input, generating a domain-specific language (DSL) that captures the OR problem, transforming the DSL into an executable script (with a suitable solver triaging), and producing a solution by executing the script. Our proposed method can significantly reduce the time and effort required to model and solve complex OR problems, which 1) reduces the time required to formulate a problem in a solver-specific format, and 2) provides a framework for the development of an interactive OR problem-solving tool that can be used by non-experts.

 We would like to emphasize that instead of providing any specific OR solution, we present a system that can create an abstract model from a user query and generate a concrete model together with the  user data and subsequently, solve the OR problem. For example in a real life scenario, a non-expert user can open a data file ( for example an Excel sheet or google sheet) and invoke(call) the system with query to solve OR problems and do "what-if" analysis. In our solution methodology, we focus on creating an abstract model grounded on a problem class instead of concrete model based on a problem instance (the problem class together with the data for a problem instance). In our methodology, the creation of an ‘abstract model’ allows users to (1) perform ‘what-if’ analysis using natural language queries, (2) deploy a problem class to support a wide spectrum of tenants with different data instances. Users can later input data to generate a ‘concrete model’ that any solver can handle. Initially, users only need access to a simple data file to create an abstract model, which they can modify to better fit different scenarios. Hence this system can be used to build out of box systems that can work with a non-technical user who wants to bring the power of AI and OR solvers to their data.

\subsection{Contributions and Organization}
Our paper distinguishes itself from existing work on the intersection of natural language processing and automated mathematical programming in the following ways:
\begin{enumerate}
    \item We are, to the best of our knowledge, the first to propose a out of the box multi-turn chat system that can generate abstract OR models from user query and then combine that with user data to generate and solve concrete OR models. Further our solution is not tied to any particular OR solver. Our approach empowers LLMs to create and edit an abstraction of a class of OR models, such as order fulfillment models. This abstraction is particularly important for industrial solution providers, as it enables them to work with the model abstraction class instead of having to work with each individual concrete OR models.
    \item  We are also the first to handle automatic solver triage. In an OR problem-solving cycle, selecting a proper solver is as important as building a mathematical program. All existing MP interfaces require developers to manually decide which solver to choose, while our approach handles solver triage through a solver interface unification. This innovation streamlines the problem-solving process and saves time for developers.
\end{enumerate}

We have implemented the proposed methodology as an end-to-end pipeline, named Natural Language to Operations Research \textit{NL2OR}, that can 1) take natural language input and generate an abstract OR model; 2) resolve data mapping and triage solvers to provide a solution to the generated OR problem; 3) edit the generated OR model for what-if analysis.

In this paper, we describe the details of our proposed method, implementation details and present experimental results on several OR problems. The rest of this paper is organized as follows. Section 2 provides an overview of related work. Section 3 describes our proposed method in detail. Section 4 presents experimental results on several OR problems. Finally, Section 5 concludes the paper and provides directions for future work.

\section{Related Works}
\subsection{Large Language Models}
A large language model (LLM)~\cite{zhao2023survey} is a sequence-to-sequence model designed for predicting sequences of text-based tokens. This model has undergone training on a substantial volume of text data to enhance its ability to generate accurate predictions for sequences of tokens in a given context. In recent years, the Transformer architecture~\cite{vaswani2017attention} has revolutionized the field by encouraging efficient parallelization of training and scalability. There have been many foundational LLM models such as BERT~\cite{devlin2018bert}, OPT~\cite{zhang2022opt}, PaLM~\cite{chowdhery2023palm}, GPT-3~\cite{brown2020language}, GPT-4~\cite{achiam2023gpt}, LLaMA~\cite{touvron2023llama}, Gemini \cite{team2023gemini}, and many more.

\emph{Prompt engineering} is a vital step in generating quality responses from a language model. A popular approach is to perform \emph{in-context learning}~\cite{brown2020language,wei2022emergent,NEURIPS2020_1457c0d6}, which is an emergent ability of LLMs where it can perform complex tasks by conditioning it on a description of the problem and few-shot examples. There are many approaches that have been shown to dramatically increase the model inference accuracy such as Chain of Thought (CoT)~\cite{wei2022chain}, Self-consistency with CoT (CoT-SC)~\cite{wang2022self}, Tree of Thoughts (ToT)~\cite{yao2024tree}, Self-consistency ~\cite{self}, ReAct ~\cite{react} , Progressive Hint Prompting ~\cite{progressive} and Graph of Thoughts (GoT) \cite{besta2023graph}.

\subsection{Automated Mathematical Programming}
An automated mathematical programming (AMP) system aims to automate the process of developing MP solutions with little of no human intervention. The first research stream of AMP focuses on developing an inference system to derive solutions based on historical data. \cite{ZHANG2023205} surveys ML-enhanced approaches that solve mix-integer programming and combinatorial optimizations. For example, researchers have found a close-box system using neural network approximations significantly improves the efficiency of modeling and solving the class of SCIP problems in the OR space. The other stream of AMP research considers the formulation of the MP problem to be articulated in natural language and subsequently translated into MP solutions utilizing LLMs \cite{LiMellou2023,ramamonjison2022,Yang2023}. Recent studies have demonstrated the efficacy of leveraging pretrained LLMs for AMP tasks~\cite{LiMellou2023,ramamonjison2022,Yang2023,prasath2023synthesis}. Due to their capacity to obviate the need for custom model training or fine-tuning, pretrained LLMs emerge as an appealing solution for AMP endeavors. Our work differentiates from existing research, as we focus on creating and editing abstract OR models, and our approach is solver-agnostic.

There have been a surge of works in the field of the optimization modelling and synthetic data generation for OR tasks \cite{optimus,chainofexp,orlm,holy,ramamonjison2022,alibaba}
following the Natural Language for Optimization (NL4Opt) NeurIPS 2022 competition. In \cite{optimus}, the authors proposed \textit{OptiMUS} a modular end to end solution taking natural language description of the problem as input and then process and solves the problem using a multi-agent framework. In \cite{alibaba}, researchers from Alibaba introduced a chat based model framework  named \textit{OptLLM} for the similar purpose. In \textit{Chain-of-Experts} \cite{chainofexp}, the model employed task specific expert agents powered by external knowledge base. A \emph{Conductor} co-ordinates  the agents effectively to create a "forward thought chain" for execution and take the feedback signals from the execution environment to trigger a "backward reflection" in order for better collaboration among the experts.

\section{Comparison with current work:}
\begin{table}
\centering
\begin{tabular}{ |c| c| c| c| c|}
\hline
Feature & \textbf{OptiGuide \cite{LiMellou2023}} & \textbf{OptiMUS \cite{optimus}} & \textbf{\makecell{Chain of\\ Experts \cite{chainofexp}}} & \textbf{\makecell{NL2OR \\ (this paper)}} \\
\hline
 \makecell{``What-if" \\ analysis} & \cmark & \xmark & \xmark & \cmark \\ 
 \hline
 \makecell{End-to-end \\ pipeline} & \xmark & \cmark & \cmark & \cmark \\ 
 \hline
 Optimizer & Gurobi & Gurobi & Gurobi  & \makecell{OR solver\\ agnostic }  \\ 
 \hline
 \makecell{External \\Knowledge} & \xmark & \xmark & \cmark & \xmark \\ 
 \hline
 \makecell{Validation,\\ Error Correction} & SafeGuard & \makecell{Evaluator \\ LLM Agent}& \makecell{Reviewer\\ LLM Agent} & \makecell{JSON Schema\\ LLM Agent}\\
 \hline
 Complexity & Low & Medium & High & Low\\ 
 \hline
 Modeling type & Concrete & Concrete & Concrete & Abstract\\ 
 \hline
\end{tabular}
\vspace{20pt}
\caption{Table comparing the functionalities of OptiGuide \cite{LiMellou2023},  OptiMUS \cite{optimus}, Chain of Experts \cite{chainofexp} with NL2OR(this paper).  }
\label{tab:Comparison}
\end{table}
In the table ~\ref{tab:Comparison}, we compare the methods and algorithms of \cite{optimus,chainofexp,LiMellou2023} and in the following, we provide a detailed discussion on each point. 
\paragraph*{What-if analysis:} By ``what-if" analysis, we mean that the constraints and the variables of the problem can change  and the algorithm or the end-to-end setup should be able  adopt the change.  The main purpose of \emph{OptiGuide:} is ``what-if" analysis. In our methodology, NL2OR can create and modify its output based on the natural language ``what-if" input.  OptiMUS \cite{optimus} and Chain-of-Experts\cite{chainofexp} do not support ``what-if" analysis.
\paragraph*{End-to-end pipeline:}    NL2OR provides ``end-to-end" architecture where the input is natural language describing the OR problem and it outputs solution alongside a report.  OptiMUS \cite{optimus} and Chain-of-Experts\cite{chainofexp} have build this capabilities but NL2OR is the only end-to-end pipeline that also supports the above mentioned "what-if" analysis. 
\paragraph*{Optimizer:}  The choice of \emph{optimizer} is important part of the end-to-end architecture for optimization based modelling. In this paper, NL2OR generates abstract model from the natural language then triaged with the appropriate solver. This is an unique capability. 
    
\paragraph*{External Knowledge:} In Chain-of-Experts \cite{chainofexp}. each experts are equipped with external knowledge base such as supply chain management scenario, Gurobi manual etc.  and facilitate In-context learning. NL2OR depends on the knowledge of the LLM.

\paragraph*{Validation, Error Correction:} All optimization model has the capability to handle error while formulating the linear program for the OR problem and generating optimization code. OptiMUS\cite{optimus} uses an ``Evaluator" that generates the error code which later being use for debugging and fixing formulation. Chain-of-Experts\cite{chainofexp} uses a ``Code reviewer" in the similar manner. In this work, we employ a custom JSON schema defining permissible properties and valid formats.

\paragraph{Complexity:} In terms of complexity, we employ a simple  methodology and architecture in comparison the agent based pipeline of  OptiMUS \cite{optimus} with a ``Manager" and Chain-of-Experts\cite{chainofexp} with ``Conductor". In addition to the "Conductor", Chain-of-Experts\cite{chainofexp} employees 11 experts (LLM) with external knowledge and reasoning with the use of Chain-of thought style prompting, summarization and comments. The methodology entails ``forward thought construction" for choosing the experts and "backward reflection" for better  collaboration among the experts. This makes the process and the system ``highly" complex. With the use of the ``Manager" and the contextual choice of agents in any step deems OptiMUS \cite{optimus} as ``medium" complexity. 

\paragraph{Modeling type:} An abstract model is a simplified representation of a system or problem, emphasizing essential elements and relationships without delving into details. It enables users to explore various scenarios and perform ‘what-if’ analyses without needing to rebuild the model. Conversely, a concrete model is a detailed and specific version of the abstract model, incorporating all necessary data and parameters, making it ready for solvers to find solutions. Our approach is novel in its focus on abstract modeling, whereas other research has primarily concentrated on concrete modeling

\begin{figure*}[!h]
\centering
\includegraphics[width=1\textwidth]{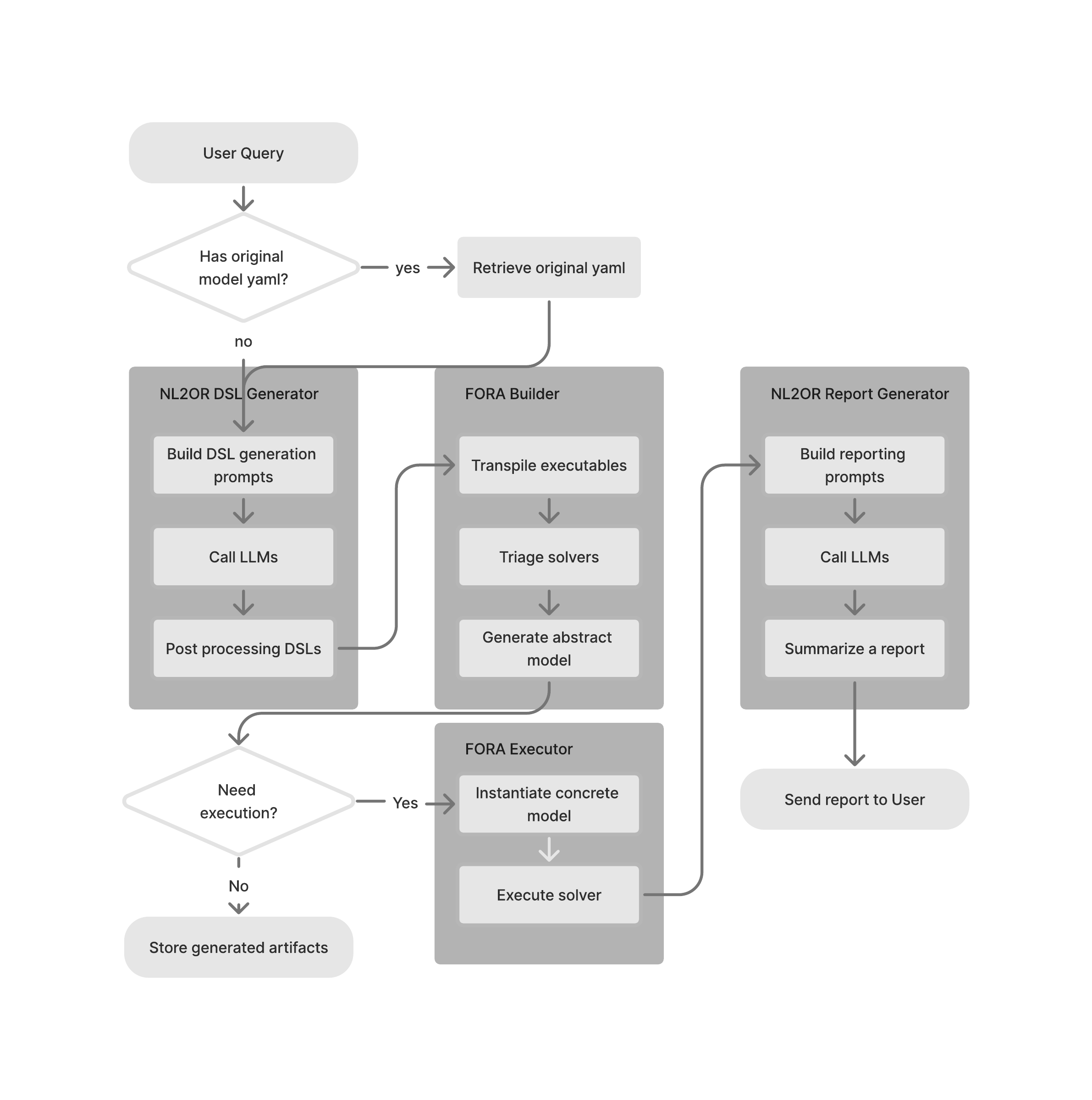}
\vspace{-20pt}
\caption{Overview of the NL2OR architecture. There are 4 major blocks in the methodology depicted in the figure. Each performs one important task. 1) With user query as input, the system first decides to create if this is a new problem or edit if it is an old problem to be updated. Then natural language input is transformed to DSL (domain specific language) which is built using prompt engineering and processed to check for error. 2) In the next step, we take the DSL and then convert it into an executable form which we call an abstract model. 3) Then with the user provided data we instantiate the concrete model and solve the problem. 4) Then, with the user query, the variable information and solution the system designs a prompt that generates a database where the solution of the problem is stored, and report is generated.}
 \label{fig:nl2or_arch}

\end{figure*}

\section{Methodology}
The main objective is of NL2OR is to convert a natural language description of an OR problem into an OR model, and then generate executable artifacts. These executables are known as abstract models, which have data input contracts. By providing data following an input contract, a concrete model instance can be solved by a selected solver. By solving the concrete model, the NL2OR pipeline can generate problem solutions with reports. This takes considerable engineering efforts to develop various parts of the pipeline. In this section, we will go through the architecture of our system in-depth. An overview of the architecture is given in Figure~\ref{fig:nl2or_arch}.

The NL2OR pipeline is composed of four major components: Domain Specific Language (DSL) Generator, Framework for OR Analytics (FORA) Builder, FORA Executor and Report Generator. The pipeline starts by admitting a user query to identify the user intention, whether the query is asking to generate a new OR model or edit an existing OR model. The DSL Generator is responsible for converting the natural language query into a formal representation of the optimization problem. The FORA Builder takes the formal representation and generates a FORA model that can be executed by the FORA Executor. The FORA Executor takes the FORA model and user specified parameter data sets, and executes it, returning an optimal solution if the concrete model is feasible. Finally, the Report Generator takes the executor's output and generates a report that can be easily understood by the user. To illustrate the pipeline, we use the example of optimizing a food purchasing plan to satisfy nutrition constraints while minimizing food purchasing cost.

\subsection{DSL Generator}
The input construct of the DSL generator variates based on a job type, i.e. creation or edition. A creation job only requires a user query, while a edition job also requires providing an original model YAML file.

Given a user input, the first step in the DSL generator is to construct a DSL generation prompt. Prompt engineering is directly interlinked with the quality of the LLM output. We streamline this process by automating prompt construction within our system. A \emph{prompt builder} module is designed to automatically generate a prompt for the LLM by amalgamating various components, including instructions, syntax overview for the (OR) model YAML, few-shot learning instances, and the user's specific problem statement. Subsequently, this constructed prompt is inputted into the LLM to facilitate the generation of the DSL, for a new OR model or an updated model.

An illustrative example of a model generation prompt is depicted in Figure \ref{fig:prompt_creation}. From the user's perspective, the complexities involved in OR model YAML and LLM prompts are abstracted away. Users must only provide a query elucidating the OR problem they aim to address. By concealing these technical intricacies, we empower users to focus solely on articulating and engaging with solutions to their OR problems without being burdened by the internal mechanics of the system. See Figure \ref{fig:prompt_creation} for an illustration of building a food purchasing planning optimization problem.

\begin{figure}[!ht]
\scriptsize

\begin{framed}
\begin{minipage}{\linewidth}
\textbf{System:}\\
You are an assistant working to generate a YAML file for an optimization process which specifies the following parameters of the process:\\

(Description of the YAML schema)\\

Refer to the examples provided with each query to determine how to generate the solution for the optimization processes.
Only output these exact outputs in the YAML file.\\

\textbf{User:}\\
Create a YAML file for an optimization process. Here is an example for the following problem:\\

Problem:\\
(Createion query example)\\

YAML file:\\
(Model YAML file example

(Additional examples...)\\

Problem:\\
InputData: food costs (vector), food nutritions (matrix), nutrition levels (vectors).\\
Variable: Decide buying quantities for each food, with different cost.\\
Objective: Minimize food purchase cost, which is a sum of production of food cost and food buying quantities.\\
Constraint: Each food contains multiple nutritions. Make sure for each category, the sum of nutrition quantities in all food bought to be no less than the minimum nutrition level, and no more than the maximum level.\\

YAML file:
\end{minipage}
\end{framed}
\caption{An overview of the LLM prompt for model creation.}\label{fig:prompt_creation}
\end{figure}
\subsubsection{Post-processing and Error Handling:}
Following the generation of the OR model YAML, i.e. the DSL, it undergoes a series of validation, correction, and processing steps facilitated by error detection mechanisms to ensure its integrity as a valid OR model. In the following, we describe the steps

\paragraph*{Syntax Error:} Initially, the YAML is subjected to \textbf{syntax error rectification}, which involves \emph{automatically correcting property names, validating correctly labeled constraints, and rectifying various Python expression errors}. These corrections aim to preempt the need for YAML regeneration if syntax errors are the only remaining issues.

\paragraph*{Early Detection:} Subsequently, beyond syntax correction, \textbf{early error detection mechanisms} are deployed to identify \emph{irreparable errors}. These mechanisms include \emph{schema validation, detection of redefined variable declarations}, and \emph{identification of undefined variables}. Upon detection of any of these errors, the system throws an error and logs it; as such, errors cannot be rectified programmatically.

\paragraph*{JSON based validation:} Validation of the YAML involves employing a custom JSON schema and defining permissible properties and valid formats for each property. For detecting redefined variable declarations, the input and decision variables are scanned to identify any duplicated names. Upon the absence of duplicates, these variables are cached as the defined set of variables, which aids in detecting undefined variables. Detection of undefined variables involves transforming Python expressions into abstract syntax trees (ASTs), enabling the extraction of non-control loop variables and subsequent verification of their prior definition as input or decision variables.
\paragraph*{Runtime Error and Restart:}
After completing these post-processing steps and confirming the absence of detectable errors, the system may still contend with the possibility of \emph{runtime errors}. Such errors may stem from either a \emph{malformed OR model} or \emph{mismatches between user data and the OR model specifications}. In cases where the OR model is malformed, regeneration of the model is necessary, necessitating a restart of the entire process. An output of the food purchase planning optimization model creation is depicted in Figure \ref{fig:or_yaml}.

\begin{figure}[!ht]
\scriptsize

\begin{framed}
\begin{minipage}{\linewidth}
\centering
\begin{lstlisting}
    InputData:
      max_nutr:
        desc: Dictionary of max nutrition values
        key:
          - nutrition: str
        value:
          - max_level: float
      min_nutr:
        desc: Dictionary of min nutrition values
        key:
          - nutrition: str
        value:
          - min_level: float
      nutr_vals:
        desc: Dictionary of nutrition values
        key:
          - food: str
          - nutrition: str
        value:
          - nutrition_value: float
      costs:
        desc: Dictionary of food costs
        key:
          - food: str
        value:
          - cost: float
    VariableBatch:
      - desc: Quantities of each food bought
        name: buy
        key:
          - food: str
        value:
          - quantity: integer
        indices: list(self.costs.keys())
        vtype: I
        lower_bound: 0
        upper_bound: inf
    Objective:
      desc: Minimize the food purchase cost
      constructor: sum(self.costs[i] * self.buy[i] \
        for i in self.costs)
      sense: min
    ConstraintBatch:
      - desc: Constraint on the total minimum nutrition
        name: min_nutr
        generator: (sum(self.nutr_vals[i, j] * self.buy[i] \ 
            for i in self.costs) >= self.min_nutr[j] \
            for j in self.min_nutr)
      - desc: Constraint on the total maximum nutrition
        name: max_nutr
        generator: (sum(self.nutr_vals[i, j] * self.buy[i] \
            for i in self.costs) <= self.max_nutr[j] \
            for j in self.max_nutr)
\end{lstlisting}
\end{minipage}
\end{framed}
\caption{Created OR model YAML for the food purchasing planning optimization problem.}\label{fig:or_yaml}
\end{figure}

A model edition prompt has the same structure with the creation prompt, with additional Original YAML file sections in few shots examples and user prompt guides. An illustrative example is depicted in Figure \ref{fig:prompt_edit}. An edition of OR model YAML file output, after DSL post processing, is illustrated in Figure \ref{fig:or_yaml_edit}. The model YAML file editions are highlighted.

\begin{figure}[!ht]
\scriptsize

\begin{framed}
\begin{minipage}{\linewidth}
\textbf{System:}\\
You are an assistant working to edit a YAML file for an optimization process which specifies the following parameters of the process:\\

(Description of the YAML schema)\\

Refer to the examples provided with each query to determine how to revise the given YAML file.
Only output these exact outputs in the YAML file.\\

\textbf{User:}\\
Based on a given query, update the original YAML, and output the updated YAML. Here are a few examples.\\

Problem:\\
(An edit query example)\\

Original YAML:\\
(Original model YAML file example)\\

Updated YAML:\\
(Edited model YAML file example)\\

(Additional examples...)\\

Problem:\\
\textit{Double the maximum nutrition levels in the model}.\\

Original YAML:
\begin{lstlisting}
    InputData:
      max_nutr:
        desc: Dictionary of max nutrition values
        key:
          - nutrition: str
        value:
          - max_level: float
      min_nutr:
        desc: Dictionary of min nutrition values
        key:
          - nutrition: str
        value:
          - min_level: float
\end{lstlisting}
\begin{center}
...
\end{center}

Updated YAML:
\end{minipage}
\end{framed}
\caption{An overview of the LLM prompt for model edition.}\label{fig:prompt_edit}
\end{figure}

\begin{figure}[!ht]
\scriptsize

\begin{framed}
\centering
\begin{lstlisting}
    InputData:
      max_nutr:
        desc: Dictionary of max nutrition values
        key:
          - nutrition: str
        value:
          - max_level: float
      ...
    ConstraintBatch:
      - desc: Constraint on the total minimum nutrition
        name: min_nutr
        generator: (sum(self.nutr_vals[i, j] * self.buy[i] \ 
            for i in self.costs) >= self.min_nutr[j] \
            for j in self.min_nutr)
      - desc: Constraint on the total maximum nutrition
        name: max_nutr
        generator:  (sum(self.nutr_vals[i, j] * self.buy[i] \
            for i in self.costs) <= 2*self.max_nutr[j] \
            for j in self.max_nutr)
\end{lstlisting}
\end{framed}
\caption{Edited OR model YAML for the food purchasing planning optimization problem.}\label{fig:or_yaml_edit}
\end{figure}

\subsection{FORA Builder and Executor}
After the LLM post-processing step, assuming successful completion, a model YAML file is transpiled into an FORA model. This transpile step primarily involves the instantiation of entities within the internal FORA library, with metadata declared in the InputData property of the YAML. The InputData property is a contractual agreement between the user and the OR model, delineating the requisite data for the solver to optimize the OR model. Consequently, it falls upon the user to furnish the specified data accordingly. Notably, the method of providing data to our system is agnostic to application specifics and, therefore, lies outside the primary scope of this work.

An illustration of this process is provided in Figure~\ref{fig:transpile_ex} for the objective. Upon loading the YAML, the objective of the OR model is instantiated by extracting the construction definition that delineates the objective. Subsequently, the associated Python expression is executed, incorporating the specified sense parameter, following which the objective is incorporated into the OR model. This procedural pattern remains consistent for the remaining properties in the OR model YAML.
\begin{figure}[ht]
\centering
\scriptsize

\begin{framed}
\begin{lstlisting}
    obj = Objective(
        constructor=exec(or_yaml["constructor"]),
        sense=or_yaml["sense"]
    )

    model.add_obj(obj)
\end{lstlisting}
\end{framed}
\caption{Example of adding the objective to an OR model.}\label{fig:transpile_ex}
\end{figure}

\begin{figure}[!h]
\scriptsize

\begin{framed}
\centering
\begin{lstlisting}
    Database Schema:
    tables:
      - name: Diet Solution
        desc: The solution report for the diet problem.
        variable: buy
        columns:
          - name: Food
            type: str
            desc: The name of the food.
            primary_key: true
            value: food
          - name: Buy
            type: int
            desc: Quantity of each food bought.
            value: quantity
\end{lstlisting}
\end{framed}
\caption{Example of a solution database schema for the diet problem.}\label{fig:report_db_schema}
\end{figure}

The generated YAML file also indicates solver specific properties, so that a generated abstract model is triaged with a proper solver during the DSL generation phase. The generated artifacts from the FORA builder is an abstract model program.

If the user furnishes input data sets, the FORA executor is then able to instantiate a concrete model. By executing the optimize method of a concrete model with the triaged solver, the FORA executor then either generates an optimal solution for feasible concrete models or returns a status flag with logs indicating infeasibility, early stop or sub-optimality.

\subsection{Report Generator}
Once FORA is employed to execute a concrete model, yielding a solution or status log, the Report Generator facilitate further analysis of the results, storing this solution in a database may be necessary. The responsibility of constructing the database schema for the solution falls under the purview of the report summarization step. The \emph{report prompt builder} plays a vital role in this process by assimilating details about the OR problem and its solution, abstracting any information related to sensitive user data, and generating a report prompt tailored for the LLM. The report prompt builder is generated using 1) the user prompt 2) variable description that is generated in the OR model and 3) the solution of the concrete model. Using the power of the LLMs, the schema is easily generated from the variable description and the column of the tables is updated with the solution values. Utilizing this prompt, the LLM generates a database schema designed explicitly for the solution to the OR model. An illustrative example of the resulting database schema is presented in Figure~\ref{fig:report_db_schema}.

After establishing this database schema, the report YAML undergoes processing, and the OR solution is transformed into an intermediate data format conducive to being seamlessly written as rows to a database. This trans-formative step prepares the OR solution data for efficient storage and retrieval in a structured database environment.

\section{Experiments}
\subsection{OR Model Creation}
For initial experimentation aimed at assessing the capabilities of NL2OR, we evaluate its proficiency in generating a valid OR model that can be executed from a specified user query. The experimentation involved assessing NL2OR across a spectrum of  30 distinct OR problem instances, spanning various scenarios encountered in practical applications. A comprehensive overview of these problem instances is presented in Table~\ref{tab:data_description}. Additionally, we explored the performance of NL2OR using two different LLM models, gpt-35-turbo-16k and gpt-4-32k, each subjected to varied temperature values. The models were evaluated based on two key metrics: \textit{Valid@k} and \textit{Latency}. Valid@k is the average number of successes generating a valid OR model when given k attempts. At the same time, Latency is the amount of time, in seconds, required for NL2OR to generate and execute the OR model. The results are given in Table~\ref{tab:results1}.

\begin{table}[!ht]
\centering
\begin{tabular}{|l|l|l|}
\hline
\textbf{Problem} & \textbf{Description} & \textbf{\# problems} \\ \hline
\begin{tabular}[c]{@{}l@{}}Minimum vertex \\ covering\end{tabular} & \begin{tabular}[c]{@{}l@{}}A vertex cover having the smallest possible\\  number of vertices for a given graph. \cite{du1998handbook}\end{tabular} & 3 \\ \hline
Knapsack optimization & \begin{tabular}[c]{@{}l@{}}Given a set of items, each with a weight and \\ a value, determine which items to  include in \\ the collection so that the total weight is less than \\ or equal to a given limit and the total value is as \\ large as possible.\cite{du1998handbook}\end{tabular} & 3 \\ \hline
\begin{tabular}[c]{@{}l@{}}Simple linear \\ programming\end{tabular} & \begin{tabular}[c]{@{}l@{}}Achieve the best outcome in a mathematical \\ model whose requirements and  objective are \\ represented by linear relationships.\cite{HillLieb01}\end{tabular} & 3 \\ \hline
\begin{tabular}[c]{@{}l@{}}Inventory assignment \\ optimization\end{tabular} & \begin{tabular}[c]{@{}l@{}}Minimize operational costs by optimizing inventory \\ assignments in a two echelon  warehouse-store network.\cite{doi:10.3138/infor.45.3.123}\end{tabular} & 3 \\ \hline
\begin{tabular}[c]{@{}l@{}}Multi echelon inventory \\ optimization\end{tabular} & \begin{tabular}[c]{@{}l@{}}Minimize supply chain costs by optimizing inventory\\ plan in a multi echelon supply  chain network.\cite{ERUGUZ2016110}\end{tabular} & 3 \\ \hline
\begin{tabular}[c]{@{}l@{}}Bill-of-material \\ optimization\end{tabular} & \begin{tabular}[c]{@{}l@{}}Maximize material utilization and product fulfillment \\ with limited raw materials in a manufacturing facility.\cite{HillLieb01}\end{tabular} & 3 \\ \hline
\begin{tabular}[c]{@{}l@{}}Network flow \\ optimization\end{tabular} & Minimize flow costs while satisfying node constraints.\cite{HillLieb01} & 3 \\ \hline
\begin{tabular}[c]{@{}l@{}}Logistic \\ optimization\end{tabular} & Minimize logistic cost in a supply chain network. \cite{rushton2022handbook}& 3 \\ \hline
\begin{tabular}[c]{@{}l@{}}Resource scheduling \\ optimization\end{tabular} & \begin{tabular}[c]{@{}l@{}}Minimize agent resource cost and maximize resource \\ utilization, while respecting  agent shift and skill \\ requirements.\cite{doi:10.3138/infor.45.3.123}\end{tabular} & 6 \\ \hline
\end{tabular}
\caption{Brief description of different optimization problems in the test dataset}
\label{tab:data_description}

\end{table}

\begin{table}[!h]
\centering
\begin{tabular}{|l|l|lll|lllll|}
\hline
\multirow{2}{*}{\textbf{Model}} & \multirow{2}{*}{\textbf{Temp}} & \multirow{2}{*}{\textbf{\makecell{Valid\\@1}}} & \multirow{2}{*}{\textbf{\makecell{Valid\\@3}}} & \multirow{2}{*}{\textbf{\makecell{Valid\\@5}}} & \multicolumn{5}{c}{\textbf{Latency (s)}}      \\ \cline{6-10}   
                       &                              &                          &                          &                          & Ave.  & Std. & P50   & P75   & P90   \\ 
\hline
gpt-35-turbo-16k & 0.1                  & 0.6              & 0.66             & 0.7            & 4.58  & 2.09 & 4.12  & 5.46  & 6.93              \\
gpt-35-turbo-16k & 0.2                  & 0.7              & 0.73             & 0.73           & 4.44  & 1.69 & 4.18  & 4.84  & 6.67                \\
gpt-35-turbo-16k & 0.4                  & 0.63             & 0.66             & 0.73           & 4.59  & 1.77 & 4.33  & 5.73  & 6.88                \\
gpt-35-turbo-16k & 0.6                  & 0.6              & 0.63             & 0.7            & 4.69  & 2.17 & 4.23  & 5.99  & 6.85                \\
gpt-35-turbo-16k & 0.8                  & 0.63             & 0.66             & 0.73           & 4.82  & 1.91 & 4.59  & 6.24  & 7.25               \\
\hline
gpt-4-32k        & 0.1                  & 0.73             & 0.7              & 0.8            & 35.49  & 15.79 & 34.17  & 45.22  & 59.34               \\
gpt-4-32k        & 0.2                  & 0.83             & 0.86             & 0.9            & 35.49  & 16.17 & 31.24  & 40.69  & 58.04               \\
gpt-4-32k        & 0.4                  & 0.63             & 0.8              & 0.9            & 36.94  & 20.40 & 32.38  & 42.53  & 55.31                \\
gpt-4-32k        & 0.6                  & 0.73             & 0.8              & 0.9            & 35.32  & 17.70 & 32.63  & 42.53  & 61.80                \\
gpt-4-32k        & 0.8                  & 0.7              & 0.73             & 0.86           & 36.77  & 15.50 & 32.99  & 39.45  & 61.59                \\
\hline

\end{tabular}

\caption{Experiments conducted on a set of 30 different OR problem creation statements}
\label{tab:results1}
\end{table}

\begin{figure*}[!h]
\centering
\includegraphics[width=1.0\textwidth]{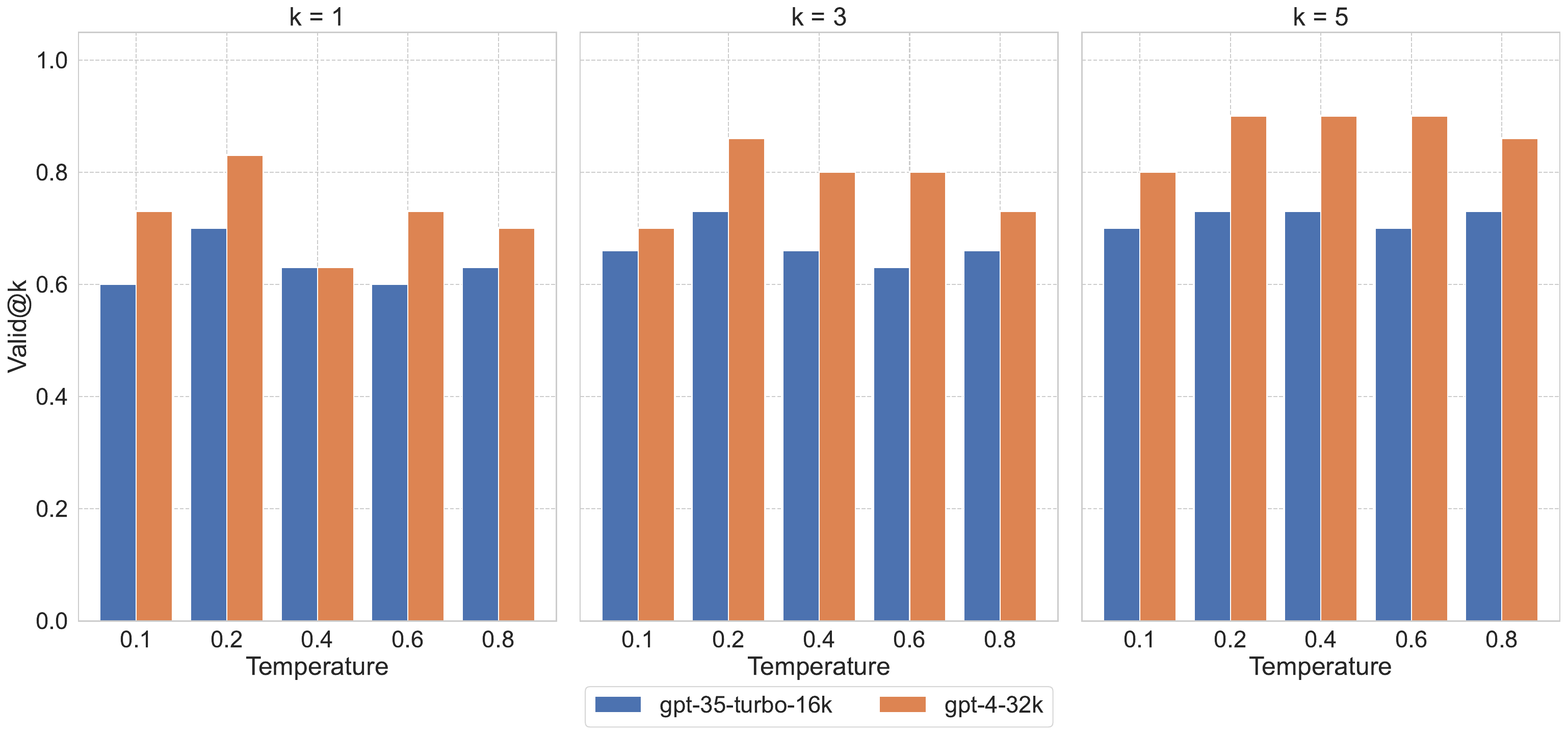}
\caption{Bar plot of NL2OR edit experiments over various k values for Valid@k and temperature values used for the LLMs.}\label{fig:edit_bar_plot}
\end{figure*}

As one might expect, gpt-4-32k has a higher latency compared to gpt-35-turbo-16k due to it having a more significant number of parameters. Additionally, we see that gpt-4-32k outperforms gpt-35-turbo-16k w.r.t. the Valid@k metric across all temperature values, showcasing its superior performance.

Both models performed similarly in producing valid YAML outputs of the OR model for well explained real world scenarios such as inventory assignment, network flow optimization or resource scheduling.
One of the major differences between the gpt-35-turbo-16k and gpt-4-32k models was in the handling of more ambiguous mathematical optimization such as optimizing the Knapsack problem or graph based algorithm solving with linear programming, where the gpt-35-turbo-16k consistently failed to produce a YAML describing a valid OR solution.

\begin{table}[!h]
\centering
\begin{tabular}{|l|l|lll|lllll|}
\hline
\multirow{2}{*}{\textbf{Model}} & \multirow{2}{*}{\textbf{Temp}} & \multirow{2}{*}{\textbf{\makecell{Valid\\@1}}} & \multirow{2}{*}{\textbf{\makecell{Valid\\@3}}} & \multirow{2}{*}{\textbf{\makecell{Valid\\@5}}} & \multicolumn{5}{c}{\textbf{Latency (s)}}      \\ \cline{6-10} 
                       &                              &                          &                          &                          & Ave.  & Std. & P50   & P75   & P90   \\ \hline
gpt-35-turbo-16k       & 0.1                          & 0.97                     & 0.97                     & 0.97                     & 4.37  & 1.75 & 3.97  & 4.42  & 8.13  \\
gpt-35-turbo-16k      & 0.2                          & 0.93                     & 0.93                     & 0.97                     & 4.37  & 1.76 & 4.02  & 4.39  & 8.08  \\
gpt-35-turbo-16k      & 0.4                          & 0.92                     & 0.92                     & 0.92                     & 4.37  & 1.75 & 3.94  & 4.40  & 8.09  \\
gpt-35-turbo-16k      & 0.6                          & 0.93                     & 0.93                     & 0.93                     & 4.38  & 1.78 & 4.03  & 4.49  & 8.09  \\
gpt-35-turbo-16k       & 0.8                          & 0.92                     & 0.93                     & 0.93                     & 4.37  & 1.77 & 3.99  & 4.49  & 8.12  \\ \hline
gpt-4-32k              & 0.1                          & 0.98                     & 1.00                     & 1.00                     & 15.00 & 7.38 & 14.19 & 15.63 & 29.46 \\
gpt-4-32k              & 0.2                          & 0.98                     & 1.00                     & 1.00                     & 15.27 & 6.66 & 13.76 & 15.54 & 28.86 \\
gpt-4-32k              & 0.4                          & 0.97                     & 1.00                     & 1.00                     & 15.46 & 6.72 & 14.04 & 15.61 & 28.47 \\
gpt-4-32k              & 0.6                          & 0.98                     & 0.98                     & 0.98                     & 15.36 & 6.82 & 13.95 & 16.07 & 28.69 \\
gpt-4-32k              & 0.8                          & 0.98                     & 0.97                     & 0.97                     & 15.24 & 6.77 & 13.76 & 15.55 & 28.77 \\ \hline
\end{tabular}
\caption{Experiments conducted on a set of 60 different OR problem edition statements}
\label{tab:results2}

\end{table}

\begin{figure*}[!h]
\centering
\includegraphics[width=1.0\textwidth]{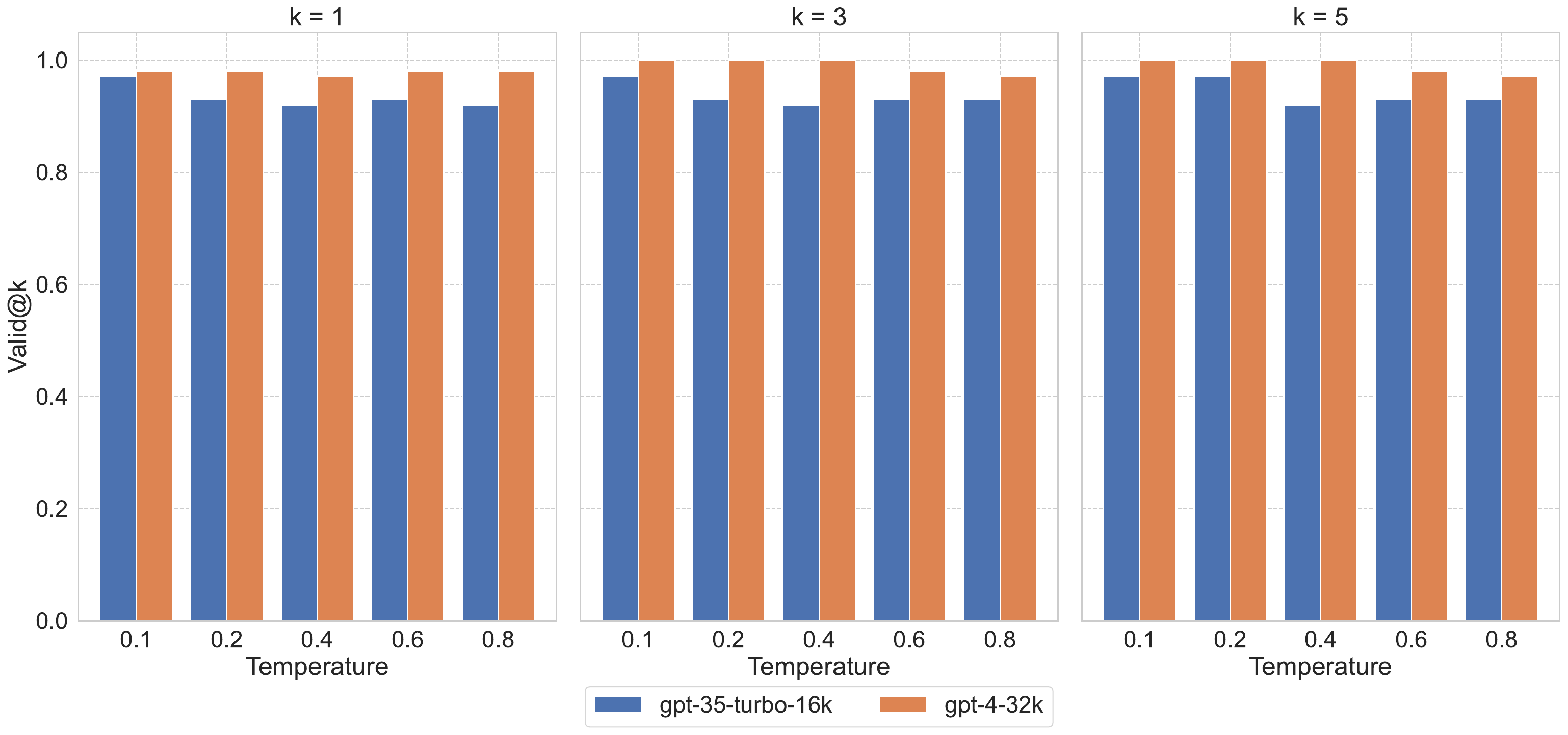}
\caption{Bar plot of NL2OR creation experiments over various k values for Valid@k and temperature values used for the LLMs.}\label{fig:create_bar_plot}
\end{figure*}

\subsection{OR Model Edit}
An essential practical application of NL2OR is editing existing OR models. We characterize OR model edits as introducing alterations or augmentations to the properties outlined in the OR model YAML. Such refinement proves invaluable when addressing minor adjustments necessary to accommodate evolving customer or business demands or validating prospective alterations to the foundational system upon which the model relies. This is also known as What-if analysis. For instance, this functionality proves invaluable when integrating a new product into a resource allocation problem or assessing the impact of updated material constraints on the resultant outcomes.

The test data is constructed as follows. We select 15 validated OR model YAML as original YAMLs, and we use GPT-4 to create possible model update queries. For each YAML file, four edition queries are generated, and thus 60 different edit scenarios are prepared. The test over 60 different edit scenarios where the results are given in Table~\ref{tab:results2}

Like the generation experiments, gpt-4-32k exhibits superior performance over gpt-35-turbo-16k concerning the Valid@k metric across all temperature settings, albeit with an associated increase in Latency. Across the entire spectrum of temperature values, NL2OR consistently demonstrates its capability to incorporate user-specified modifications into the OR model accurately. A comparative analysis with the creation experiments underscores the relative simplicity of modifying an existing OR model compared to generating a novel one. Intuitively, this makes sense, as translating user intentions into components of an OR model is much easier when the prior underlying system is given.

\subsection{OR Model Creation for LPWP Data:}
We evaluate the OR Model creation problem with the LPWP dataset \cite{ramamonjison2022augmenting} that is collected from the NL4Opt competition in NeurIPS 2022. We evaluate NL2OR Model creation on $287$ test samples data with three different LLM models, gpt-35-turbo-16k, gpt-4-32k and the latest model gpt-4o. We choose the temperature to be $0.1$ for all the experiments. In the table \ref{tab:results3}, we show the accuracy (\emph{Valid$@1$)}, average number of prompt tokens for each LLMs, the total cost to run the experiments and latency. Accuracy is the number of successful generation of the OR models out of $287$ problems. We compute the prompt tokens from the \emph{usage} of the chat completion api of each LLM models. We calculate the cost in the following way,
\begin{align*}
    \text{Total Cost} &= (\text{Prompt Tokens} / 1000000) * \text{Cost per 1M input tokens}) \\
    &+   (\text{Completion Tokens} / 1000000) * \text{Cost per 1M output tokens})
\end{align*}

The pricing for all the models can be found \cite{pricing}. All the models performed really well. As expected, gpt-4o performed the best in terms of accuracy $(94.4\%)$ and the cost is lowest for gpt-35-turbo-16k. For the gpt-4 the accuracy is close to the gpt-4o but the cost higher than that of gpt-4o model. This is due to the fact that gpt-4o is designed to be more cost effective than gpt-4 which is a very large model. In terms of latency, gpt-35-turbo-16k has the lowest latency and gpt-4-32k being a large model has the highest latency. gpt-4o performed the best in terms of accuracy, latency and cost overall.
\begin{table}[!h]
\centering
\begin{tabular}{|l|l|l|l|lllll|}
\hline
\multirow{2}{*}{\textbf{Model}} & \multirow{2}{*}{\textbf{\makecell{Accuracy\\(Valid@1)}}}   & \multirow{2}{*}{\textbf{\makecell{Prompt Token\\(Mean $\pm$ std)}}} & \multirow{2}{*}{\textbf{\makecell{Total Cost\\(USD)}}} &  \multicolumn{5}{c}{\textbf{Latency (s)}}      \\ \cline{5-9} 
                    &    &                      &                          & Ave.  & Std. & P50   & P75   & P90   \\ \hline
gpt-4o & $94.4\%$ & $2230.76\pm 22.12$ & $2.84$ &  5.01 & 2.01  & 4.49 & 5.47 & 6.93\\
\hline
gpt-4-32k & $91.6\%$ & $2852.61\pm 22.34$ & $30.45$  & 7.22 & 1.48 & 7.08 & 8.0 &  9.08\\
\hline
\makecell{gpt-35\\-turbo-16k} & $77.3\%$ & $2852.61\pm 22.34$ & $0.57$ & 3.93& 0.93 & 3.8 & 4.43&  4.968\\
\hline
\end{tabular}
\vspace{5pt}
\caption{Experimental result on LPWP Dataset.}
\label{tab:results3}
\end{table}
\color{black}
\section{Conclusion and Future Work}
In this paper, we propose a novel methodology that creates new OR solutions and edits existing OR solutions using natural language (NL) queries. Our proposed method can significantly reduce the time and effort required to model and solve complex OR problems, which reduces the time required to formulate a problem in a solver-specific format and provides a framework for the development of an interactive OR problem-solving tool that can be used by non-experts. We have implemented the proposed methodology as an end-to-end pipeline, named \textit{NL2OR}, that can 1) take natural language input and generate an abstract OR model; 2) resolve data mapping and triage solvers to provide a solution to the generated OR problem; 3) edit the generated OR model for what-if analysis. To demonstrate experimentally, first, we evaluate NL2OR across a spectrum of more than 30 distinct OR problem instances, spanning various scenarios encountered in practical applications where NL2OR shown to have high accuracy and low latency.  Additionally, we show how NL2OR is capable of effectively editing existing OR models. Finally, we choose the LPWP dataset \cite{ramamonjison2022augmenting} and show the effectiveness of NL2OR in terms of accuracy and cost across $(\approx) 300$ test problems. It is always beneficial to use smaller models as they are faster and cheaper. In our experiment, we show the cost and accuracy for different LLMs. A comprehensive study into the use and effectiveness of  SLM is an important future work. Moreover, a study on the use of reinforcement learning to improve the performance of the NL2OR is an exciting direction.

\bibliographystyle{splncs04}
\bibliography{references}

\end{document}